\documentclass[10pt,english,journal,a4paper]{IEEEtran}

\usepackage{soul}
\usepackage{url}
\usepackage[hidelinks,bookmarks=false]{hyperref}
\usepackage[utf8]{inputenc}
\usepackage[font=small]{caption}
\usepackage{graphicx}
\usepackage{amsmath}
\usepackage[ruled,vlined]{algorithm2e}
\usepackage{subcaption}
\usepackage{enumerate}
\usepackage{amsfonts}
\usepackage[numbers]{natbib}
\usepackage[utf8]{inputenc}
\usepackage[english]{babel}
\usepackage{amsthm}
\usepackage{booktabs}
\usepackage{tabularx}
\usepackage[colorinlistoftodos]{todonotes}
\usepackage{multirow}

\theoremstyle{definition}

\title{Hierarchical Program-Triggered Reinforcement Learning Agents For Automated Driving}

\author{
 Briti Gangopadhyay \and Harshit Soora \and
Pallab Dasgupta*\thanks{*The authors would like to thank DST, Govt of India, and TCS Research Scholarship for partial financial support of this project.}\\
 Department of Computer Science and Engineering, IIT Kharagpur, India\\
 \{briti\_gangopadhyay,soora\}@iitkgp.ac.in, pallab@cse.iitkgp.ac.in
}

\begin{document}

\maketitle

\begin{abstract}
Recent advances in Reinforcement Learning (RL) combined with Deep Learning (DL) have demonstrated impressive performance in complex tasks, including autonomous driving~\cite{DRLsurvey}. The use of RL agents in autonomous driving leads to a smooth human-like driving experience, but the limited interpretability of Deep Reinforcement Learning (DRL) creates a verification and certification bottleneck. Instead of relying on RL agents to learn complex tasks, we propose HPRL - Hierarchical Program-triggered Reinforcement Learning, which uses a hierarchy consisting of a structured program along with multiple RL agents, each trained to perform a relatively simple task. The focus of verification shifts to the master program under simple guarantees from the RL agents, leading to a significantly more interpretable and verifiable implementation as compared to a complex RL agent. 
The evaluation of the framework is demonstrated on different driving tasks, and NHTSA pre-crash scenarios using CARLA, an open-source dynamic urban simulation environment. 
\end{abstract}

\section{Introduction}
There has been a steady increase in the development of self-driving cars as they possess the potential to radically change the future of mobility. Deriving safe driving policies remains a key challenge in achieving deployable autonomous driving systems. The formulation of driving strategies has been studied by three schools of work, namely rule-based methods, imitation based learning, and reinforcement learning.

Rule-based methods for behaviour and motion planning have been studied and developed for decades. They rely on symbolic computation techniques such as finite state machines as used by \cite{darpa1,darpa2} in the DARPA challenges or classical planning algorithms \cite{glaser,karla}. However, rule-based methods require the driving problem to be modelled in terms of domain-specific rules in an underlying logical language. This often is not feasible for autonomous driving where the environment is dynamic as well as stochastic and can give rise to an enormous number of driving scenarios. The manual encoding of logical rules also has high human involvement. 

To minimize the human component in the formulation of driving policies, imitation based learning techniques have been explored \cite{cnn1,cnn2}. The goal of imitation learning is to mimic driving behaviour from data that has been extracted from human driving in a supervised fashion. The key idea is to map sensor information like camera images to some indicators that can directly be translated into control values such as throttle and steering angle by fitting a function on available driving data \cite{deepdrive}. However, imitation learning has a high reliance on a huge amount of annotated driving data, is challenging to scale, and the control policies learnt are only as good as the data available. 

Reinforcement learning (RL) eliminates the requirement of prior knowledge and labelled data as it learns policies by trial and error while trying to optimize a cumulative future reward function. These algorithms have displayed superhuman performance in games like GO \cite{alphago}. Classical reinforcement learning techniques, both model-based and model-free such as Q learning cannot be adopted directly to solve complex domains such as autonomous driving as they have large state/action space. Hence, deep learning techniques have played a vital role where an approximation of the Q function or the RL policy is learned implicitly by a neural network-based architecture. RL techniques leveraging deep learning have been successfully used to learn stochastic policies in simulation environment like TORCS~\cite{QLearning,mnih}. 

Driving is a sequence of complex manoeuvres which often leads to the problem of receiving sparse rewards when it is modelled solely as a RL problem. Deep learning techniques are known to suffer from a cold start and require extensive amounts of training for converging to reasonable policy. Also, because of the non-interpretable and opaque nature of neural networks, functional safety, which is a vital requirement for certification of safety-critical systems, cannot be guaranteed~\cite{shashua}.

\begin{figure*}[!t]
\begin{tabular}[b]{c@{\hskip 1.2in}c}
    \begin{subfigure}[b]{0.50\linewidth}
        \includegraphics[scale=0.45]{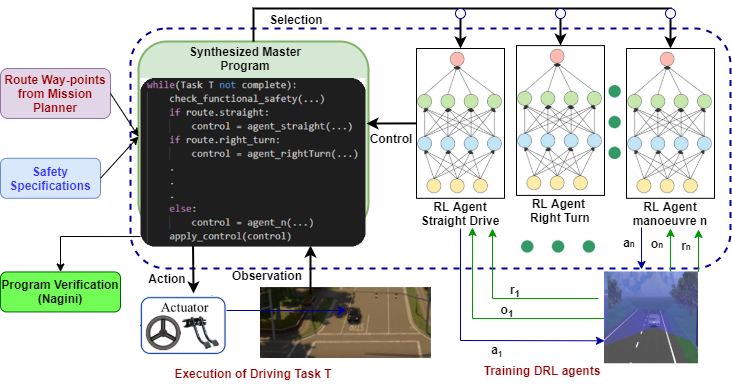}
        \caption{}
        \label{fig1:a}
    \end{subfigure}
    &
    \begin{subfigure}[b]{0.30\linewidth}
            \includegraphics[scale=0.40]{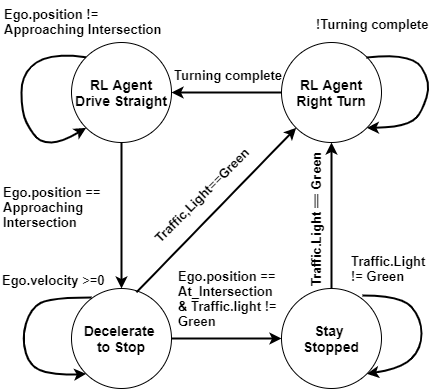}
            \caption{}
            \label{fig1:b}
    \end{subfigure}
\end{tabular}
\caption{(a) The Hierarchical Program triggered Reinforcement Learning (HPRL) Framework. The right side of the figure depicts the asynchronous training of RL agents with individual observation/action space and reward signal. The left part depicts the program-controlled execution of a driving task where the program checks for functional safety and triggers RL agents as per route plan generated by the mission planner or as per control strategy recommended by safety module. (b) State machine for right turn manoeuvre at a 4-way traffic light controlled intersection without any dynamic objects as described in example \ref{fourway}}
\end{figure*}

Human beings often use a delicate combination of traffic/road rules and experience while driving. The human driving policies are semantic rather than being geometric and numerical, for example, ``drive straight till the junction and turn right" rather than ``drive 50 meters at the current speed and steer $90^o$"~\cite{rss}. These semantic instructions have sub-goals which can be contoured into hierarchical abstractions. The sub-goals, in turn, are heavily regulated by symbolic rules that are applied at various points of time and space (for example, stop at the red light). Inspired by this, we propose a Hierarchical Program triggered Reinforcement Learning (HPRL) framework for executing autonomous driving tasks. We asynchronously train reinforcement learning agents for learning different driving manoeuvres using manoeuvre specific actions, observation space and reward signals, making them more sample efficient than a single flat DRL policy. The agents are then triggered in execution by a symbolic rule-based system interpreted in terms of a structured program to complete a driving task. Functional safety requirements are incorporated as embedded assertions in the program, thereby facilitating a formal proof of the correctness of the driving strategy. The framework has been tested on NHTSA pre-crash scenarios in CARLA \cite{carla}. 

\noindent In summary, this work makes the following novel contributions:
\begin{itemize}

    \item Training a set of generic manoeuvres containing the behaviours {\em drive\_straight, right\_turn, left\_turn, change\_left\_lane, change\_right\_lane}, using model-free value-based Deep Q Learning (DQN) and policy optimization-based Deep Deterministic Policy Gradient (DDPG) networks with manoeuvre specific constrained action and state spaces and reward signals.
    
    \item Decomposing driving tasks in terms of the generic manoeuvres and controlling them using a structured Program $\mathcal{P}$ to satisfy the overall driving intent while shielding the RL agents with respect to the safety specifications $\varphi$. The correctness of $\mathcal{P}$ with respect to $\varphi$ is verified using the python program verification tool Nagini \cite{nagini}.
    
\end{itemize}
The paper is organized as follows. Section ~\ref{sec2} discusses related work. Section~\ref{sec3} presents the preliminaries and the problem statement, Section~\ref{sec4} presents the design methodology, and Section~\ref{sec5} presents details on the verification methodology and experimental work. Section~\ref{sec6} provides concluding remarks.

\section{Related Work}
\label{sec2}
The growth of deep reinforcement learning has given traction to the development of self-driving policies in the absence of labelled data. However, the non-interpretable and non-explainable nature of deep learning techniques do not make them suitable for use in safety-critical domains. As a result, the neuro-symbolic research community has been working towards merging the classical symbolic computation based techniques with deep learning, making them more sample efficient and interpretable. We explore each of these relevant areas of research in more details.

\subsection{Reinforcement Learning in Autonomous Driving} The competitive and cooperative nature of driving tasks makes model-free reinforcement learning a suitable choice for learning driving policies as the model is not fixed apriori. In~\cite{Guan2018} it has been shown how a Markov Decision Process (MDP) formulation can be used for self-driving cars in a highway with state discretization. Several prior works attempt to learn a complete policy directly from perception and sensor data~\cite{endtoend,deepframework,TacticalDM,learningtodrive}. This approach is both computationally expensive and time-consuming. Recently there has been a shift from learning entire policies using vanilla deep reinforcement learning, to decomposing the policies into sub-policies and learning them via Hierarchical Deep Reinforcement Learning \cite{hierrl}. The use of an {\em option graph} in a multi-agent setting has been explored in~\cite{shashua} and tested on a lane negotiation scenario. In this case, the reinforcement learning agents are only trained to learn desires such as comfort. In~\cite{TowardsPH,Duan2020}, a hierarchical reinforcement learning framework has been proposed for multi-lane autonomous driving, where both behaviour and motion planners are trained using deep RL, but this does not enable us to provide any guarantees of functional safety. {\cite{hrlLanechange} Use Hierarchical DRL to learn sub-policies related to lane-change with temporal and spatial attention to image data, while \cite{zeroshot} uses HRL in trajectory planning.} The RL agents in these works are limited to the manoeuvres, {\em switch lane left}, {\em switch lane right}, and {\em keep lane}. \\
{ DRL, though extensively used to approximate policies for complex systems like autonomous drive in simulation, are impeded in practical control applications for well-known drawbacks like sample complexity, sensitivity to hyper-parameters, network architecture and reward signal, interpretability and safety \cite{concrete}. Our work investigates whether we can retain equivalent performance, as compared to using a single complex RL policy, by breaking the policy into multiple sub-policies and training individual RL agents for each of them facilitating sample efficiency. A deterministic supervisory controller designed as a structured program triggers RL agents during execution. This enables us in establishing safety guarantees over the structured program rather than the RL policies.} 

\subsection{Symbolic computation and Reinforcement Learning} An important direction of modern AI is to find synergistic combinations of classical AI, which uses symbolic reasoning, and connection-based AI, which uses deep neural networks. Hierarchical reinforcement learning guided exploration using AI planning has been studied in~\cite{darling,peorl,sdrl}. Planning guided reinforcement learning drastically reduces the exploration space, thereby handling the problem of cold start. The planning framework requires the environment to be predefined in terms of logical rules. A recent study \cite{IntegratingDR} leverages information from path planners by integrating distance to closest way-point as a part of RL state space. Our present contribution is most closely related to~\cite{prl}, where given a task in a natural language, ambiguities are resolved by a structured formal program which triggers a single agent to fulfil the task. On the other hand, we decompose driving tasks into a set of sub-goals and train different agents for each of these goals, and therefore our use case and our approach is very different from that of~\cite{prl}. Moreover, we demonstrate the verifiability of our designs by translating functional safety requirements into assertions over the structured program. Our framework works on a continuous space discrete/continuous action problem using a driving simulator.

\section{Problem Statement}
\label{sec3}
A finite horizon discounted Markov Decision Process (MDP) is defined over the tuple $\langle \cal{S,A,}$$P_{ss^{'}}^{a}$$\cal{,R,\gamma} \rangle$ where:
\begin{itemize}
    \item $\cal{S}$ is the set of states, 
    \item $\cal{A}$ denotes the entire set of low-level actions an agent can perform,
    \item $P_{ss^{'}}^{a}$ denotes the probability of transition from state $s \in \cal{S}$ to state $s^{'} \in \cal{S}$ by taking action $a \in \cal{A}$. This transition probability is not explicitly defined in model free RL. 
    \item $\cal{R}$ denotes the reward signal.
    \item $\gamma$, $0 \leq \gamma \leq 1$ is the discount factor.
\end{itemize}
Given an MDP, the goal of Reinforcement Learning (RL) is to find a policy $\pi: \cal{S} \to \cal{A}$ that maximizes the reward signal over the finite horizon. In Hierarchical Reinforcement Learning (HRL), a task is solved by decomposing it into a series of sub-goals using temporal abstractions. In~\cite{barto} options with three components are considered, namely a policy $\pi : \cal{S} \times \cal{A} \to $ [0,1], a termination condition $\beta : S \to $ [0,1], and an initiation set $\cal{I} \subseteq S$. An option $(\cal{I},\pi, \beta)$ is available in state $s_t$ iff $s_t \in I$. In our framework, the initiation set and the terminating conditions of an option is decided by a program $\cal{P}$. $\cal{P}$ triggers RL agents trained for different generic driving manoeuvres to perform the overall task. The lower level RL agents orchestrate with each other via the meta controller program to satisfy the global intent of the driving task.

Our aim is to develop a verifiable framework where we decompose a driving task $\cal{T}$ into a set of sub-tasks $t \in \cal{T}$. For each sub-task, a set of RL agents $\cal{X}$ is pre-trained. Each $x_t$ is only exposed to a subset of relevant actions $a_t \in \cal{A}$ and trained using reward signal $r_t$, so that the training is sample-efficient. Program $\mathcal{P}$ triggers policy $\pi_t$ of agent $x_t \in X$ with initialisation states $\mathcal{I}_t$ and terminating condition $\beta_t$ depending on the way-points generated by the mission planner. $\mathcal{P}$ also has embedded safety specifications $\varphi$ for shielding the agents $\cal{X}$ from taking unsafe actions. Agents from $\cal{X}$ are triggered until task $\cal{T}$ is completed (Fig.~\ref{fig1:a} shows the framework).

\section{Methodology Overview}
\label{sec4}

We provide the methodology overview using a running example. Section~\ref{fourway} outlines a simple driving task as an example. Section~\ref{rlagents} defines the state space, action restrictions and reward function for behaviour specific DRL agents. Section~\ref{forspec} develops the formal safety specification, and Section~\ref{embed} outlines the verification methodology.

\subsection{Four-Way Intersection Scenario} \label{fourway}
We consider a simple scenario of a straight drive and right turn on a 4-way traffic light controlled intersection. We refer to the subject vehicle, namely the one controlled by HPRL as the {\em ego vehicle}. The driving task, $\cal{T}$, is to drive the ego vehicle from location A to location B, which involves the following sub-tasks as defined in natural language:
\begin{enumerate}
    \item Sub-Task $t_1$ : The ego vehicle should drive straight while maintaining lane and desired speed till the junction.
    \item Sub-Task $t_2$ : The ego vehicle must stop on observing a red light and remain stationary until the light turns green.
    \item Sub-Task $t_3$ : The ego vehicle should take a right turn to reach the destination (for this manoeuvre).
\end{enumerate}
The desired behaviour can be represented as the state machine in Fig.~\ref{fig1:b}. A program $\cal{P}$ derived according to the state machine first triggers the RL agent $x_1$ that controls the straight driving of the ego vehicle while tracking speed and maintaining lane for sub-task $t_1$. The ego vehicle decelerates as it approaches the intersection. When the ego vehicle stops at the intersection $\cal{P}$ checks the traffic light and keeps the vehicle stationary until the light turns green. This behaviour is according to the rules of the road and does not require learning. Once the light turns green $\cal{P}$ initiates the right turning RL agent $x_2$ and completes sub-task $t_3$, thereby completing the global driving task $\cal{T}$. An inherent advantage of using a hybrid hierarchical framework like this is that certain safety properties can be formally asserted, such as {\em the ego vehicle must remain stationary at a red light until the light turns green}. This kind of functional safety guarantees cannot be easily obtained in flat or hierarchical RL policies.

\subsection {Reinforcement Learning Agents} 
\label{rlagents}
The proposed HPRL platform maintains a library of simple manoeuvres and corresponding pre-trained RL agents. For example, let us consider the following set of manoeuvres:
\begin{itemize}
    \item {\em Agent drive\_straight}. Track speed, keep lane, and maintain specified safe distance from leading vehicle.
    \item {\em Agent right\_turn}. Negotiate a right turn. More details to follow.
    \item {\em Agent left\_turn}. Similar to Agent right\_turn.
    \item {\em Agent change\_left\_lane}. Responsible for making a safe transition to the lane on the left of the present lane. More details to follow.
    \item {\em Agent change\_right\_lane}. Similar to Agent { change\_left\_lane}.
\end{itemize}
For each of these manoeuvres, we train a value-based Deep Q Learning (DQN) agent and policy-based DDPG agent in the CARLA simulator. These agents can have overlapping behaviours like lane-keeping, keeping safe distance etc.  For the policy network $\theta^\mu$, which suggests actions that maximize expected reward, and objective function $J(\theta)$, the policy gradient update is calculated as:
\begin{equation}
\begin{aligned}
& \nabla_{\theta^\mu}J_i(\theta) =  \left [ \mathbb{E}_{(s,a,r)} \left ( \nabla_{\theta^\mu}\mu(s_i|\theta^\mu)\nabla_aQ(s_i,a_i|\theta^q)  \right ) \right ]
\end{aligned}
\end{equation}
The loss for each DQN agent and critic network for DDPG agent is calculated using the following loss function at each iteration, $i$:
\begin{equation}
\begin{aligned}
& L_i(\theta) = \\
& \left [ \mathbb{E}_{(s,a,r)} \left ( r+\gamma \mathop{max}_{a_{t+1}}Q^{\theta^{q-}_{i}}(s_{t+1},a_{t+1})-Q^{\theta^q_{i}}(s_t,a_t)  \right )^2 \right ]
\end{aligned}
\end{equation}
Q-Learning updates are applied on $\langle state,action,reward \rangle$ samples by drawing random samples from the data batch. $\theta^q_i$ represents the Q/critic-network parameters and $\theta^{q-}_i$ are the target network parameters at iteration i.

We distribute the reward over a continuous display of correct behaviour and provide a positive reward $r_{sub\_goal}$ when the episode terminates correctly by achieving the sub-goal. For example, for the straight driving agent maintaining lane and keeping a safe distance is a display of correct behaviour and will fetch positive reward at each time step and $r_{subgoal}$ will be added on reaching the target way-point. The state space, allowed actions and reward signal for each of the manoeuvres are as follows:
\vspace{0.2in}

\subsubsection{Straight Driving Agent}
The objective of this agent is to track speed, keep lane and maintain a safe distance from leading vehicles. The observation space ${\cal S}_{straight}$ is ($\delta x_t$,$\delta x_c$,$v_{ego}$,$\delta x_{lon}$) where $\delta x_t$ is the L$_{2}$ norm distance from the sub-task target way-point, $\delta x_c$ is the distance from the centre of the lane, $v_{ego}$ is the current speed of the ego vehicle and $\delta x_{lon}$ is the distance from the leading vehicle. The action space ${\cal A}_{straight}$ can control acceleration, deceleration and restricted steering. The reward function $r_{straight}$ is a combination of the following rewards:

\begin{equation}
\left\{
\begin{tabular}{ll}
$r_{c} = -C_1$ & Collision\\ 
$r_{l} = -1*(\delta x_c)$ & \text{Diff\_Center\_Lane}\\ 
$r_{v} = \beta*(v_{target}-v_{ego})$ &  Diff\_Target\_Speed  \\ 
$r_{lon} =  1*(\delta x_{lon})$ & Diff\_Lead\_Vehicle \\
$r_{t} = -1*(\delta x_t)$ & Diff\_Target\_Location \\
\end{tabular}
\right\}
\end{equation}

\noindent
$C_1$ is a constant. $\beta = -1$ if $v_{target} \geq v_{ego}$, else $\beta = 1$. 
\[ {r}_{straight} = r_{c}+r_{l}+r_{v}+r_{lon}+r_{t}+r_{sub\_goal} \]

\subsubsection{Right / Left Turning Agents} 
These agents execute the turning instructions. The observation space for $\mathcal{S}_{right}$ and $\mathcal{S}_{left}$ is ($\delta x_c$, $\delta x_t$, $\delta x_n$, $v_{ego}$, $\delta_{angle}$) where $\delta x_n$ denotes the distance from the nearest object and $\delta_{angle}$ refers to the difference between the ego vehicle's headway angle and target way-point angle. The action space $\mathcal{A}_{right}$ and $\mathcal{A}_{left}$ contain restricted acceleration. $\mathcal{A}_{right}$ has access to only right steering actions, and $\mathcal{A}_{left}$ has access to only left steering actions. The reward function  $r_{right}$ and $r_{left}$ are defined as follows:
\begin{equation}
\left\{
\begin{tabular}{ll}
    $r_{c} = -C_1$ & Collision\\ 
    $r_{l} = -1*(\delta x_c)$ & Diff\_Center\_Lane\\  
    $r_{n} =  1*(\delta x_{n})$ & Diff\_Nearest\_Object\\
    $r_{a} = -1*(\delta_{angle})$ & Diff\_Angle \\
    $r_{t} = -1*(\delta x_t)$ & Diff\_Target\_Location\\
\end{tabular}
\right\}
\end{equation}
\[ {r}_{left/right} = r_{c}+r_{l}+r_{n}+r_{a}+r_{t}+r_{sub\_goal} \]

\subsubsection{Change Left / Right Lane Agents} 
The left/right lane change agents can be triggered based on the proposed path by the mission planner. They can also be triggered if the path of the ego vehicle is blocked beyond a particular time threshold, $t_\epsilon$, and if the clearance distance, $d_\epsilon$, is available for lane change. This helps in avoiding static obstacles and slow leading vehicles. The  observation  space $S_{left\_change}$ / $S_{right\_change}$ is defined over the tuple ($yaw_{ego}$, $\delta_{yaw}$, $\delta x_t$, $v_{ego}$), where $yaw_{ego}$ is the motion about the perpendicular axes, and $\delta_{yaw}$ is the difference between the yaw of the way-point and the vehicle. When $\delta_{yaw}$ = 0, the vehicle is perfectly aligned with the road. The action spaces $\mathcal{A}_{right\_change}$ and $\mathcal{A}_{left\_change}$ contain restricted acceleration and restricted left and right steering. The reward function, $r_{right\_change}$ / $r_{left\_change}$ is as follows:
\begin{equation}
\left\{
\begin{tabular}{ll}
    $r_{c} = -C_1$ & Collision\\ 
    $r_{ch} = -C_2$ & Incorrect\_Lane\\  
    $r_{st} =  -C_3*(steer)$ & Incorrect\_Steer\\
    $r_{yaw} = -1*(\delta_{yaw})$ & Diff\_Yaw \\
\end{tabular}
\right\}
\end{equation}
$C_2, C3$ are constants. The reward $r_{ch}$ applies when the ego vehicle is not driving in the desired lane. $r_{st}$ effects the reward when there is an undesired steering. For example if the left lane change is underway and the ego vehicle has reached the desired lane but is still steering towards left, then $r_{st}$ adds to the reward. $r_{yaw}$ accounts for the difference between the targeted yaw angle and the current yaw angle. 
\[ r_{left/right change} = r_{c}+r_{ch}+r_{st}+r_{yaw}+r_{sub\_goal} \]

\begin{figure*}[!t]
\begin{tabular}[b]{cr}
\begin{subfigure}[b]{0.50\linewidth}
        \includegraphics[scale=0.33]{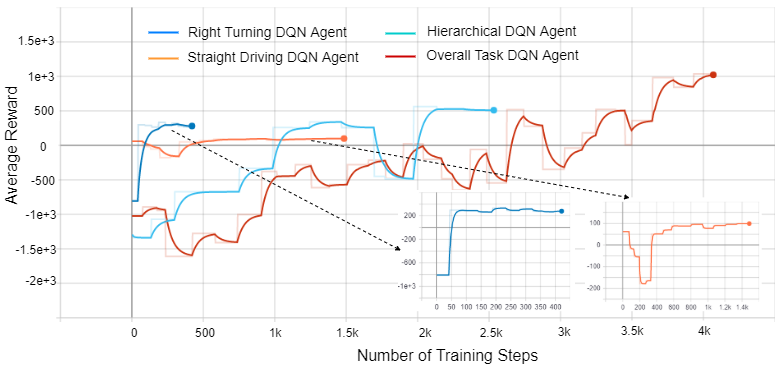}
        \caption{}
        \label{fig2:a}
    \end{subfigure}
    &
    \begin{subfigure}[b]{0.50\linewidth}
        \includegraphics[scale=0.35]{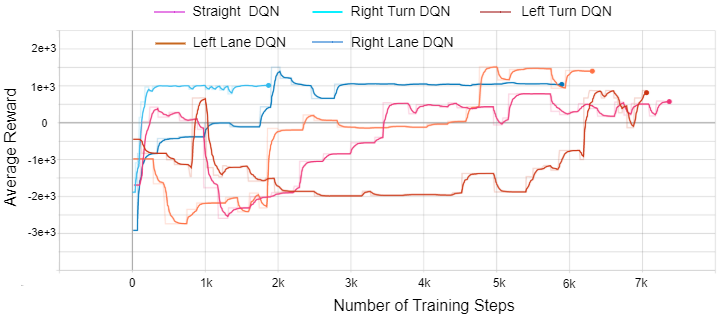}
        \caption{}
        \label{fig2:b}
    \end{subfigure}
    \\
    \begin{subfigure}[b]{0.50\linewidth}
        \includegraphics[scale=0.48]{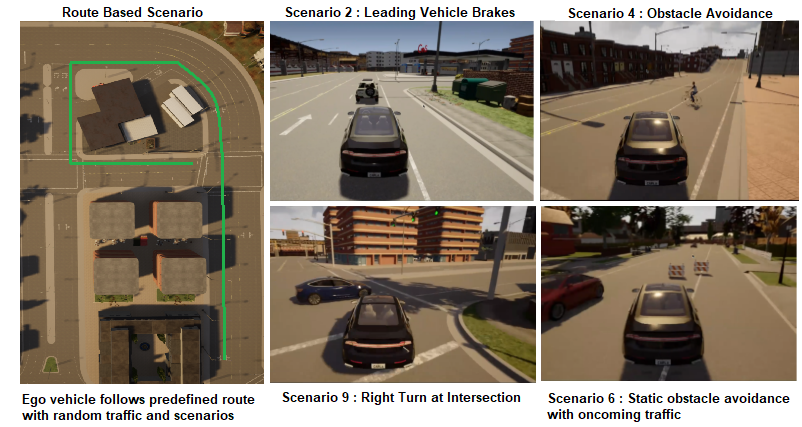}
        \caption{}
        \label{fig2:c}
    \end{subfigure}
    &
    \begin{subfigure}[b]{0.45\linewidth}
        \includegraphics[scale=0.33]{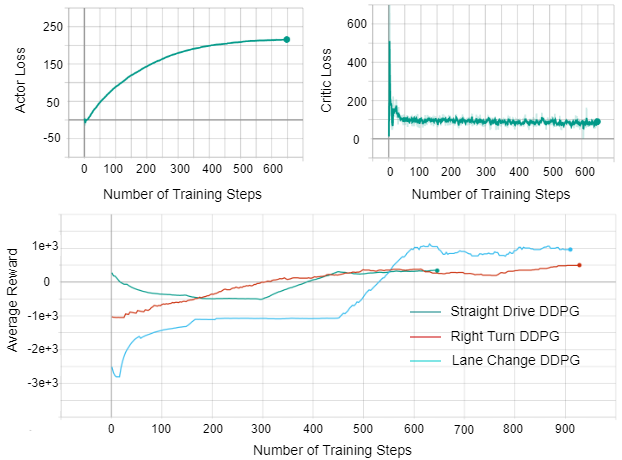}
        \caption{}
        \label{fig2:d}
    \end{subfigure}
    \end{tabular}
    \caption{(a) Comparison between individual agents learning separate policies, Hierarchical DQN choosing between two policies and flat DQN learning the overall policy for task \ref{fourway}. (b) Average Rewards per step for all manoeuvre specific DQN models (c)  HPRL agent tested on scenarios inspired by NHTSA pre-crash scenarios in CARLA { and route based tasks with random traffic and scenarios.} The scenario descriptions are at Table at \ref{tab:2} and simulation video links are available \cite{hprlvideos} (d) Actor Loss, Critic Loss for DDPG straight driving model and Average Reward for DDPG straight, right turn and left lane change models}
\end{figure*}

\subsection{Safety Specifications} \label{forspec}
The structured program $\cal{P}$, besides acting as the sequencer and trigger for the RL agents, also acts as a safety shield developed from safety specifications gleaned from well known driving rules. These kinds of safety specifications have been studied \cite{rss,stl,sff}. {In symbolic model checking instead of enumerating reachable states of a state machine with embedded entry exit conditions one at a time, the state machine can be efficiently checked by taking a cross product with automaton constructed from the safety property ($B\ddot{u}chi$ automaton). $B\ddot{u}chi$ automaton is an automata-theoretic version of a formula in Linear Temporal Logic \cite{Pnueli1977TheTL} which can encode temporal safety specifications \cite{lttobuchi}. Temporal logic extends classical propositional
logic with a set of temporal operators that navigate between
a set of time steps. For brevity we use LTL for describing the safety specifications rather than showing its $b\ddot{u}chi$ automata equivalent.} The syntax of LTL is given by the following grammar:
\[
\phi := \textbf{T} \:|\:\neg\: \phi \: |\:\phi_1\:\lor\:\phi_2|\:\bigcirc\phi\:|\:\lozenge\:\phi\:|\:\square\:\phi\:|\:\phi_1 \:\mathcal{U}\: \phi_2
\]
where \textbf{T} is true, $\bigcirc$ is the next operator { (the property should hold in the next time step)}, $\lozenge$ is the future operator {(the property should hold eventually sometime in the future)}, $\square$ is the global operator {(the property should hold at all time steps)} and $\cal{U}$ is the until operator { ($\phi_1\cal{U}$$\phi_2$ means $\phi_1$ should hold until the time step where $\phi_2$ becomes true)}. We also use the release operator $\cal{R}$ where $\phi_1 \, \mathcal{R} \, \phi_2 \equiv$  $\neg(\neg\phi_1 \, \mathcal{U} \, \neg\phi_2)$. The semantics of LTL can be found in \cite{Katoen}.

The Release operator $\phi_1 \, \mathcal{R} \, \phi_2$  means that $\phi_2$ always holds up to the time when $\phi_1$ becomes true. We use a modified version of the release operator as proposed in \cite{stl}, $\overline{\cal{R}}$, which does not require $\phi_2$ to hold at all if $\phi_1$ has occurred in the past. We shall first specify the functional safety requirements in a restricted fragment of LTL, and then demonstrate the translation of this specification to embedded assertions in the structured program. Essentially, the functional safety module in $\cal{P}$ checks and triggers emergency responses overriding the current RL policies if any of the following properties hold. 
\begin{enumerate}
    
    \item {\em The ego vehicle shall stop if the traffic light is red, and remain stationary until the traffic light is not red.} Suppose $e$ denotes the ego vehicle, proposition $e_{stop}^{lat}$ represents no lateral movement, and $e_{stop}^{lon}$ represents no longitudinal movement of the ego vehicle. We define:
    \[ e_{stop} = e_{stop}^{lat} \, \land \, e_{stop}^{lon} \]
    {The proposition $T_{red}$ is set to true when the traffic light is red. We want to encode the property that if the traffic light is not red at time step $t_1$ and changes to red at time step $t_2$ then from time step $t_2$ the ego vehicle should start applying break and come at a stop both laterally and longitudinally till the stop manoeuvre is released by the lights changing to not red at some time step $t_r$. The property should hold at all time steps and hence is wrapped by global operator}. The property can then be coded in LTL as:
    \[ \square(\neg T_{red} \land \bigcirc T_{red} \to \bigcirc(\neg T_{red} \; \overline{\mathcal{R}} \; e_{stop})) \]
    
    \item \label{assertion2}{\em The ego vehicle shall not make any longitudinal movement if its distance with the lead vehicle falls below a specified safety threshold}. Let the proposition, $\textit{D}_{e,v}^{lon}$, be true when the longitudinal safe distance between ego vehicle $e$ and leading vehicle $v$ is safe, and false otherwise. The proposition, $\textit{L}_e$, becomes true when lane change is triggered for $e$. The property can be coded in LTL as:
    \[ \square(\textit{D}_{e,v}^{lon} \land \neg \bigcirc  \textit{D}_{e,v}^{lon} \to \bigcirc((\textit{D}_{e,v}^{lon} \lor \textit{L}_e) \; \overline{\mathcal{R}} \;e_{stop}^{lon})) \]
    It be noted that the restriction is removed when the leading vehicle has moved forward to a safe distance or a when lane change is triggered.
    
    \item {\em The ego vehicle shall not make any lateral movement towards a neighboring vehicle if the lateral distance between the two vehicles fall below a specified safety threshold}. Suppose the proposition, $\textit{D}_{e,v}^{lat}$, be true when the lateral distance between the ego vehicle $e$ and the lateral neighbor $v$ is safe, and false otherwise. The proposition, $\textit{L}_{e/v}$, is true when a lane change is triggered for $e$, but not towards the lane occupied by $v$. The property can be coded in LTL as:
    \[ \square(\textit{D}_{e,v}^{lat} \land \neg \bigcirc  \textit{D}_{e,v}^{lat} \to \bigcirc((\textit{D}_{e,v}^{lat} \lor \textit{L}_{e/v}) \; \overline{\mathcal{R}} \; e_{stop}^{lat})) \]
    
    \item  {\em At a junction, the ego vehicle must remain stationary until the vehicles of higher priority have cleared the junction. The priority of a vehicle is higher if the vehicle has entered the junction earlier. Also, if the routes $r_1$ of ego vehicle and $r_2$ of vehicle $v$ intersect, then the vehicle having a smaller distance to the set $r_1 \cap \, r_2$ has higher priority.} Suppose the proposition, $\textit{J}_{e,v}$, is true when the ego vehicle and another vehicle, $v$, have both entered the route to junction. The safety requirement may be specified in LTL as:
    \[ \square(\textit{J}_{e,v} \land (\textit{Priority}(v) > \textit{Priority}(e)) \to \neg J_{e,v} \; \overline{\mathcal{R}} \;e_{stop}) \]
    
    \item {\em If the ego vehicle has initiated a lane change manoeuvre, $\textit{L}_{e}$, and the target lane does not have a clearance of at least $\textit{C}_{\epsilon}$ required for the lane change operation then the ego vehicle remains stationary until the target lane is clear}. We may express this requirement in LTL as:
    \[ \square(\textit{L}_{e} \land \neg \textit{C}_{\epsilon} \to \textit{C}_{\epsilon} \; \overline{\mathcal{R}} \;e_{stop}) \]
\end{enumerate}

{Position estimation and tracking of vehicles as well as vulnerable road users such as pedestrian and bicycles can be done using any standard methods, such as use of object localization neural networks and sensors such as radar/lidar, stated in \cite{cycledetect}. In our experiments we do position estimation via sensors such as available in CARLA and occupancy grid map. While changing lane we take  $\textit{C}_{\epsilon}$ to be the length of the car with a 2m space in front and rear. For an opposite lane manoeuvre $\textit{C}_{\epsilon}$ is a function of the time taken for the ego vehicle to overtake the static/dynamic obstacle and velocity of the incoming traffic.}

\subsection{From Safety Specification to Embedded Assertions} \label{embed}
Though there exists a rich arsenal of tools for formal verification of software, and an equally rich arsenal of tools for LTL model checking, there are no mature offerings for checking LTL properties on C, C++, or Python. One direction pursued by some researchers is to translate the program into languages accepted by LTL model checking tools, such as Promela (for using SPIN, LTSmin) \cite{SPIN,LTSmin}, or SMV (for using NuSMV) \cite{NuSMV}. The main challenge here is to establish that the translation is semantically correct and correctly models the behavior with respect to the truth of the specified LTL properties.

An alternative approach, which we choose to follow in this work, is to translate the LTL specification into a form that can be formally verified using program verification tools. Since our code is developed in Python, we choose Nagini \cite{nagini} as our verification platform. {Nagini is an automatic verification tool, based on the Viper verification infrastructure, for statically typed Python programs.} In this subsection, we outline the methodology for translating the LTL properties into embedded assertions in Nagini. This is not an easy task in general, but the HPRL framework has a structure which makes this possible, as discussed in this section. {Our aim is to formally prove that the safety shield always guards against any failure of the formal properties specified earlier. We illustrate how this is done using one of the specified properties.}

A high-level view of our structured program is shown in Algorithm \ref{algo:1}. As described before, the role of the structured program is to {take a sequence of sub-routes for a driving task as input and to invoke suitable (pre-trained) RL agents for executing each sub-route. The sub-route division from source to destination is obtained by CARLA's inbuilt path planner GlobalRoutePlanner (A class which can be used to dynamically compute trajectories from an origin to target waypoints).} Since the vehicle operates in a dynamical environment, the structured program must examine the state of the system at periodic intervals to determine whether the scenario permits the present RL agent to continue, or whether a new agent needs to be invoked. For example, it may choose to replace the {\em straight driving agent} by a {\em lane change agent} if the lead vehicle stops or slows down in the present lane. In order to implement this behavior, the structured program uses the function, {\em neural\_control(RouteList)}, to invoke the suitable RL agent for the current state and current sub-route in the {\em RouteList}. The return value is the actuation recommended by the RL agent, that is, values of throttle, steering, etc.

In order to ensure safe execution, we build a safety wrapper around the RL agent. Instead of seeking the control actuation values directly from the function, {\em neural\_control(RouteList)}, our structured program calls the function, {\em check\_functional\_safety(RouteList)}, which uses a safety shield over the RL agents. This function evaluates the state of the system and determines whether it is safe to allow the RL agent to recommend the next actuation. If so it calls {\em neural\_control(RouteList)} and returns the actuation values computed by it; otherwise, it overrides the RL agent and seeks safe actuation values from a safe controller via the function, {\em safe\_control\_$\varphi$(~)}. {The safe controller in our context is the high level program responsible for providing set points to an actual underlying controller such as the braking control. The actuation values are determined based on the safety specification for which safe\_control\_$\varphi$(~) is overriding the RL agent currently in execution. For example, if the ego vehicle encounters a red light \ref{forspec} the safe actuation set by safe\_control\_$\varphi$(~) is to start applying brake and no steering and throttle (setting brake = 1.0, steering = 0.0 and throttle = 0.0 in CARLA).}

\begin{algorithm}[t]
\SetAlgoLined
\DontPrintSemicolon
\caption{ \small Structured Program $\mathcal{P}$}
\label{algo:1}
\KwIn{$Ego Vehicle$, $RouteList$, $Environment$}
\SetKwFunction{FMain}{Main}
\SetKwFunction{FSafe}{check\_functional\_safety}
\SetKwProg{Fn}{Function}{:}{}
\Fn{\FMain{$RouteList$}}{
    \While{True}{
        $control =$ check\_functional\_safety($RouteList$)\;
        $Ego Vehicle.Apply(control)$\;
        $Time.sleep(
        \delta t)$
    }
 }
 \vspace{0.2cm}
 \Fn{\FSafe{$RouteList$}}{
    \For{each $\varphi$ in $specifications$}{
         \If{$\neg \varphi$}{
            $control$ = $safe\_control\_\varphi()$\;
            \KwRet{control}\;
        }
    }
    $control$ = $neural\_control(RouteList)$\;
    \KwRet{control}\;
 }
\end{algorithm}

Nagini requires input programs to comply with the static, nominal type system defined in {PEP 484 \footnote{https://www.python.org/dev/peps/pep-0484/} (standard syntax for function annotations in python)}. Hence, the modules of $\cal{P}$ that we intend to verify are converted to their statically typed equivalents and annotated with assertions. For statically typed concurrent python programs, Nagini is capable of proving  memory safety, data race freedom, and user-supplied assertions~\cite{nagini}. Assertions in Nagini are provided in the form of $Assert(Implies(e_1, a_2))$ which denotes that assertion $a_2$ holds if Boolean expression $e_1$ is true. We consider the validation of Specification~\ref{assertion2} discussed in section \ref{forspec} to elucidate the conversion from LTL specification to embedded assertions in Nagini specification language. Specification~\ref{assertion2}, is expressed as: 
\[ \varphi = \textit{D}_{e,v}^{lon} \land \neg \bigcirc  \textit{D}_{e,v}^{lon} \to \bigcirc((\textit{D}_{e,v}^{lon} \lor \textit{L}_e) \; \overline{\mathcal{R}} \;e_{stop}^{lon}) \] 
The following methods are annotated with assertions:
\begin{table*}[t]
\small
\begin{tabular}{p{5cm}p{12cm}}
\toprule
Architecture  $\theta^{q-}, \theta^q$    & DQN : (State Space) X 64 X 32 X (Action Space), 

DDPG : Concatenate(Radar Space,State Space) X 256 X 128 X 64  X 32 X (Action Space)
\\ \midrule
Architecture $\theta^{\mu^-}, \theta^{\mu}$ & Radar : (Radar Space) X 512 X 256, Odometry: (State Space) X 64 X 32 X 16,

Total Input : Concatenate(Radar,Odometry,272) X 256 X 128 X 32 X (Action Space)\\
\midrule
$\epsilon$-greedy parameters & $\epsilon_0 = 1$, $\lambda_{decay} = 0.995$, $\epsilon_{min} = 0.03$ \\ \midrule
Straight Driving Agent & Throttle$_{range}$ = 0.2\dots1,
Steer$_{range}$ = -0.1\dots0.1\\ \midrule
Right/Left Turning Agents &  Throttle$_{range}$ = 0.3\dots0.6, Steer$_{right}$ = 0.2\dots0.5, Steer$_{left}$ = -0.2\dots-0.5\\ \midrule         Right/Left Lane Change Agents & Throttle$_{range}$ = 0.4\dots0.5, Steer$_{right}$ = 0.1\dots0.3,  Steer$_{left}$ = -0.1\dots-0.3\\
\bottomrule
\end{tabular}
\caption{Architecture, Hyper-parameters and action range for the DQN and DDPG agents. The throttle value ranges from 0 to +1 and steering value ranges from -1 to +1 which have been restricted for different agents. The action space has been discretized for DQN Agents and is continuous for DDPG Agents.}
\label{tab:1}
\end{table*}
\begin{enumerate}

    \item Method {\tt \_is\_vehicle\_hazard(\dots)} invoked from within the function, \textit{check\_functional\_safety()}, is responsible for checking whether there is a violation of longitudinal safety constraint with all detected leading vehicles. This method inspects if the $L_2$ distance between the leading vehicles and the ego vehicle is hazardous, and if so, sets the variable \textit{long\_hazard\_detected} to {\tt True} and returns {\tt True}.
    
    In order to formally prove the assertion, $\varphi$, we need to first prove that the method {\tt \_is\_vehicle\_hazard(\dots)} returns {\tt True} when the $L_2$ distance is less than the specified safe distance, that is, {\em norm\_distance $\leq$ proximity\_threshold}. This is formally proven using the following assertion in Nagini.
    
    def \_is\_vehicle\_hazard(vehicle\_List  :\\ List[Vehicle],ego\_vehicle :  Vehicle)$\to$ bool:\\
      ...
      Assert(Implies(norm\_distance $\leq$\\ proximity\_threshold,Result()==True))
      
    This assertion guarantees that this method sets the variable \textit{long\_hazard\_detected} when the antecedent of $\varphi$ is true, namely $\textit{D}_{e,v}^{lon} \land \neg \bigcirc  \textit{D}_{e,v}^{lon}$.

    \item Inside the \textit{check\_functional\_safety(~)} method, the \textit{emergency\_stop(~)} method applies full brake and keeps applying brake, when \textit{long\_hazard\_detected} is {\tt True}, until the leading vehicle is beyond unsafe distance or a lane change is triggered. This should cause ego vehicle brake to take a value of 1.0 which implies full brake in CARLA environment and steering and throttle value to be 0.0 which is the required behaviour $e_{stop}^{lon}$. The assertion verifying \textit{long\_hazard\_detected} leads to vehicle.brake == 1.0 and accounts for $\textit{D}_{e,v}^{lon} \land \neg \bigcirc  \textit{D}_{e,v}^{lon} \to  \;e_{stop}^{lon}$. 
    
    Also, as part of the implementation the ego vehicle initiates a lane change by setting \textit{lane\_change} = True, which is equivalent to setting the truth value of $\textit{L}_e$, if ego vehicle remains stopped beyond blocking time $\ge$ $t_{\epsilon}$  (\textit{self.\_blocking\_threshold}). This behaviour is ensured by the assertion : {\em if self.time $\ge$ blocking\_threshold then lane\_change is set to {\tt True}}.

def check\_functional\_safety(self) $\to$ Control:\\
  long\_hazard\_detected = self.\_is\_vehicle\_hazard(vehicle\_list)\\
  if long\_hazard\_detected and not lane\_change:\\
    self.time\_count =  self.time\_count + 1\\
    if self.time\_count $>$ self.\_blocking\_threshold:\\
      lane\_change = True\\
  if long\_hazard\_detected and not lane\_change:\\
    control = self.emergency\_stop()\\
  Assert(Implies((self.time\_count $>$\\
  self.\_blocking\_threshold and not lane\_change), lane\_change))\\
  Assert(Implies((long\_hazard\_detected and not lane\_change),\\
  self.vehicle.brake == 1.0 and\\
  self.vehicle.steer == 0.0 and\\ 
  self.vehicle.throttle == 0.0))\\

    \item Finally we want to ensure $(\textit{D}_{e,v}^{lon} \lor \textit{L}_e) \; \overline{\mathcal{R}} \;e_{stop}^{lon}$. If longitudinal hazard is not detected and lane change is not underway we validate that \textit{check\_functional\_safety()} returns the control suggested by triggering neural networks via the function \textit{execute\_nn\_control()} which changes the $e_{stop}^{lon}$ behaviour. If a lane change is underway then \textit{lane\_change} is {\tt True} and this triggers the function \textit{execute\_lane\_change()} which also releases the $e_{stop}^{lon}$ behaviour.

while True:\\
  control = agent.check\_functional\_safety()
  world.ego\_vehicle.apply\_control(control)

def check\_functional\_safety(self) $\to$ Control:\\
  ...
  Assert(Implies((not(long\_hazard\_detected) and not(lane\_change)),\\
  control == self.execute\_nn\_control()))\\
  Assert(Implies lane\_change,\\
  control == self.execute\_lane\_change()))\\
  return control 
\end{enumerate}
\vspace{0.5cm}
In general given a safety specification in the form of
\[ \varphi = t_0p_0t_1p_1 \dots t_kp_k \to t_ka_kt_{k+1}a_k \dots t_rp_r\]
where $t_0\dots t_r$ are time steps, $p_0 \dots p_k$ are the truth values of a set of propositions at each time step, $t_r, p_r$ are the release time and release proposition respectively and $a_k$ is an action, the validation task is to validate the following.
\begin{enumerate}
    \item For the antecedent: Validate that the proposition set $p_i$ is set to {\tt True} if the conditions for the propositions are satisfied at a given time step $t_i$. For example, combinations of many evaluation factors can lead the condition \textit{proximity.threshold $\leq$ norm\_distance} to be {\tt True} in the method \textit{\_is\_vehicle\_hazard()}. However, the validation module only needs to validate whether the corresponding predicate \textit{long\_hazard\_detected} is set to {\tt True} if the condition holds at any time step.
    \item For the consequent:  Validate if the antecedent is satisfied at time step $t_k$ then the recommended action $a_k$ is triggered from $t_k$ till set of propositions $p_r$ are {\tt True} at release time $t_r$. For example, we validate the ego-vehicle remains at a stop till it is released by lane change or safe longitudinal distance. 
\end{enumerate}

\section{Empirical Studies}
\label{sec5}
The implementation of our framework consists of two parts, as shown in Fig. \ref{fig1:a}. The first part is the implementation of RL agents using  DQN and DDPG networks which use state, action space and the reward functions introduced in section \ref{rlagents}. The architectural details of the networks used are mentioned in Table \ref{tab:1}. The simulator CARLA is used for both training and validation of HPRL agents. To evaluate the sample efficiency of sub-DQN agents, with $\mathcal{P}$ as a trigger, we compare them to a flat DQN agent and a Hierarchical DQN agent where all of them try to learn the task described in Example \ref{fourway} as shown in Fig \ref{fig2:a}. The  Hierarchical DQN learns to choose between the straight DQN and the Right Turn DQN to accomplish the task. The HPRL framework can directly trigger the straight driving and right turning agents to complete the task. Both the Hierarchical DQN and Flat DQN take more training steps to achieve the task, and neither of them can provide functional safety.

The average reward per 100 training steps obtained by all the DQN sub-agents on complete training (learning generic manoeuvres) is shown in Fig \ref{fig2:b}.
The
trade-off between exploration and exploitation while training DQN sub-agents is
handled by following an $\epsilon$-greedy policy. We use a replay buffer of length 1000000 and a discount factor of $\gamma = 0.99$. We also train our agents using Policy optimization-based DDPG networks to achieve action over a continuous action space. The DDPG network uses an extra input of radar cloud-points along with the states. The average reward obtained by the DDPG sub-agents at each training step is illustrated in Fig \ref{fig2:d}. The training and testing of all the experiments are performed on a machine with 6-core 2.4 GHz Intel Core i7 8th Gen and 4GB NVIDIA GeForce GTX TITAN.\\
We validate the second part of our implementation $\mathcal{P}$ with Nagini. Nagini takes 50.71s to validate all assertions for LTL specifications in $\mathcal{P}$, using Z3 SMT solver in the back-end, which generates 18,477 clauses.\\
We test the performance of HPRL framework on 10 NHTSA inspired pre-crash scenarios modelled in CARLA\footnote{\url{https://carlachallenge.org/challenge/nhtsa/}}. The description of each scenario, along with the RL agents and assertions triggered in each case are described Table \ref{tab:2}, and pictorial representation is shown is Figure \ref{fig2:c}. The implementation code is open-sourced and available at \cite{hprl}. The framework is also tested on longer driving tasks where the ego vehicle travels on a predefined route. We spawn 120 traffic vehicles which randomly move throughout the town and trigger random scenarios on the ego vehicle's path.

\begin{table*}[t]
\small
\begin{tabular}{p{3.5cm}p{7.5cm}p{3.8cm}p{1.5cm}}
\toprule
\textbf{Scenario} & \textbf{Expected Behaviour} & \textbf{DRL Agents} & \textbf{Assertions}\\ \midrule
Control Loss & Ego vehicle loses control without prior action and must recover & Straight Drive & None\\ \midrule
Longitudinal control & Leading vehicle decelerates suddenly and ego-vehicle must perform an emergency brake or an avoidance maneuver.  & Straight Drive, Lane Change & 2,5\\ \midrule
Obstacle avoidance with/ without prior action & While performing a maneuver, the ego-vehicle finds an obstacle must perform an emergency brake or an avoidance maneuver.  & Straight Drive, Lane Change & 2,3,5\\ \midrule
Lane change  & Ego-vehicle performs a lane changing to evade a leading vehicle, which is moving too slowly.  & Straight Drive, Lane Change & 2,3,5\\ \midrule
Vehicle Passing with oncoming traffic & Ego-vehicle must go around a blocking object using the opposite lane, yielding to oncoming traffic & Straight Drive, Lane Change & 2,3,5\\
\midrule
Running Red Light at Intersection & Ego-vehicle is going straight at an intersection but a crossing vehicle runs a red light, ego-vehicle must perform a collision avoidance maneuver & Straight Drive & 1,2,4\\
\midrule
Unprotected left turn at intersection  & Ego-vehicle is performing an unprotected left turn at an intersection, yielding to oncoming traffic. & Straight Drive, Left Turn & 1,3,4\\
\midrule
Right turn at an intersection  & Ego-vehicle is performing a right turn at an intersection, yielding to crossing traffic & Straight Drive, Right Turn & 1,3,4\\
\midrule
Crossing at an un-signalized intersection  & Ego-vehicle needs to negotiate with other vehicles to cross an un-signalized intersection with the first to enter the intersection having priority.  & Straight Drive & 2,4\\
\bottomrule
\end{tabular}
\caption{Description of the test scenarios along with the triggered agents and assertions. The simulation videos are available at \cite{hprlvideos}}
\label{tab:2}
\end{table*}

\section{Conclusions and Future Work}
In this paper, we present the Hierarchical Program Triggered Reinforcement Learning (HPRL) framework, which uses deep reinforcement learning agents triggered  by a structured program embedded with rule-based safety specifications. The experiments demonstrate that DRL agents trained with manoeuvre specific actions and state spaces are sample efficient. We show that the framework is capable of handling challenging pre-crash scenarios for autonomous driving vehicles. The framework integrates formal validation with the program verification tool Nagini to ensure functional safety. In future,
the authors would like to reduce the DRL agent's granularity, for example, to braking and steering, to facilitate continuous switching between them making the control smoother. The authors also wish to study how safety specifications can directly be embedded as a part of the DRL agents, thereby eliminating the requirement of an explicit safety shield.

\label{sec6}
\def\bibindent{0.5em}
\bibliographystyle{unsrt}
{\small\bibliography{ijcai20}}
\vspace{-1cm}
\begin{IEEEbiography}[{\includegraphics[width=1in,height=1.25in,clip,keepaspectratio]{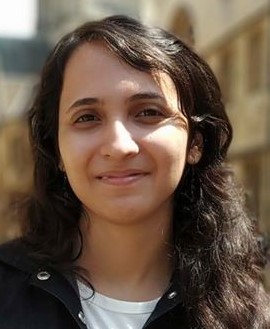}}] {Briti Gangopadhyay} (Student Member, IEEE) is a research scholar at department of computer science and engineering, IIT Kharagpur, Kharagpur, India. She is a part of the Formal Methods and Trusted AI Group, IIT Kharagpur.  Her current research areas include Explainable Artificial Intelligence, Verifiable Autonomous driving policies, and Neuro-Symbolic Reasoning. She is also recipient of TCS Research Fellowship.
\end{IEEEbiography}
\vspace{-1cm}
\begin{IEEEbiography}[{\includegraphics[width=1in,height=1.25in,clip,keepaspectratio]{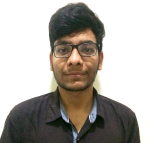}}] {Harshit Soora} is a final year undergraduate student in the department of computer science and engineering, IIT Kharagpur, Kharagpur, India. He is currently a part of Trusted AI Group. His current research areas include reliable autonomous policies and verifiable deep reinforcement learning.
\end{IEEEbiography}
\vspace{-1cm}
\begin{IEEEbiography}[{\includegraphics[width=1in,height=1.25in,clip,keepaspectratio]{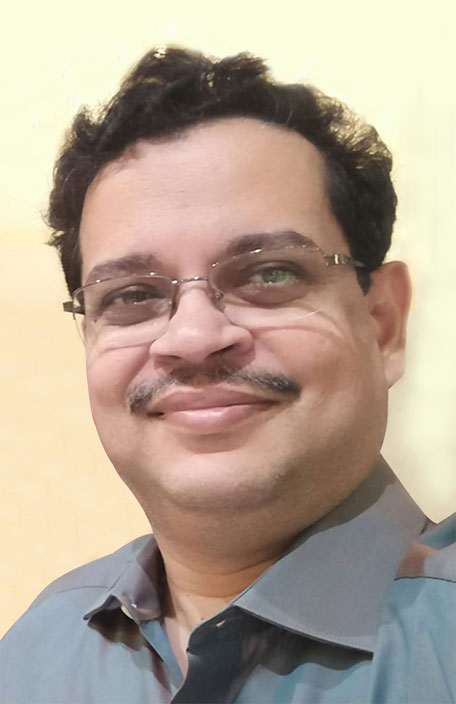}}] {Prof.\ Pallab Dasgupta} (Senior Member, IEEE) received the Ph.D. degree in computer science and engineering from IIT Kharagpur, in 1995. He is currently a Professor in Computer Science and Engineering with IIT Kharagpur. He leads the Formal Methods and Trusted AI Group with collaborations at Intel, Synopsys, SRC, Texas Instruments, Indian Railways, and HAL. He has more than 200 research papers. Dr. Dasgupta is a Fellow of the Indian National Academy of Engineering and the Indian Academy of Science.
\end{IEEEbiography}

\end{document}